\documentclass[11pt,a4paper]{article}
\usepackage[hyperref]{emnlp2020}
\usepackage{times}
\usepackage{latexsym}

\usepackage{microtype}

\usepackage{multirow}
\usepackage{booktabs}

\aclfinalcopy 

\newcommand*{\enpl}{En~$\rightarrow$~Pl}
\newcommand*{\plen}{Pl~$\rightarrow$~En}

\title{Tilde at WMT 2020: News Task Systems}

\author{Rihards Krišlauks$^{\dagger \ddagger}$ \and Mārcis Pinnis$^{\dagger \ddagger}$  \\
  $^\dagger$Tilde / Vienibas gatve 75A, Riga, Latvia \\
  $^\ddagger$Faculty of Computing, University of Latvia / Raiņa bulv.  19, Riga, Latvia\\ 
  {\tt \{firstname.lastname\}@tilde.lv} \\}

\date{}

\begin{document}
\maketitle
\begin{abstract}
This paper describes Tilde's submission to the WMT2020 shared task on news translation for both directions of the English$\leftrightarrow$Polish language pair in both the constrained and the unconstrained tracks. 
We follow our submissions form the previous years and build our baseline systems to be morphologically motivated sub-word unit-based Transformer base models that we train using the Marian machine translation toolkit. Additionally, we experiment with different parallel and monolingual data selection schemes, as well as sampled back-translation.  Our final models are ensembles of Transformer base and Transformer big models which feature right-to-left re-ranking.
\end{abstract}

\section{Introduction}
This year, we developed both constrained and unconstrained NMT systems for the English$\leftrightarrow$Polish language pair. We base our methods on the submissions of the previous years \cite{pinnis-EtAl:2017:WMT,Pinnis2018news,pinnis-krilauks-rikters:2019:WMT} including methods for parallel data filtering from \newcite{Pinnis2018}.
Specifically, we lean on \newcite{Pinnis2018} and \newcite{junczys2018dual} for data selection and filtering, \cite{pinnis-EtAl:2017:WMT} for morphologically motivated sub-word units and synthetic data generation, \newcite{edunov_understanding_2018} for sampled back-translation and finally \newcite{morishita_ntts_2018} for re-ranking with right-to-left models. We use the Marian toolkit
\cite{mariannmt} to train models of Transformer architecture
\cite{vaswani2017attention}. 

Although document level NMT as showcased by \cite{junczys2019microsoft} have yielded promising results for the English-German language pair, we were not able to collect sufficient document level data for the English-Polish language pair. As a result, all our systems this year translate individual sentences. 

The paper is further structured as follows: Section~\ref{sec:data} describes the data used to train our NMT systems, Section~\ref{sec:exp} describes our efforts to identify the best-performing recipes for training of our final systems, Section~\ref{sec:res} summarises the results of our final systems, and Section~\ref{sec:conclusion} concludes the paper.


\section{Data}
\label{sec:data}
For training of the constrained NMT systems, we used data from the WMT~2020 shared task on news translation\footnote{http://www.statmt.org/wmt20/translation-task.html}. For unconstrained systems, we used data from the Tilde Data Library\footnote{https://www.tilde.com/products-and-services/data-library}. The 10 largest publicly available datasets that were used to train the unconstrained systems were Open Subtitles from the Opus corpus \cite{tiedemann2016finding}, ParaCrawl \cite{banon2020paracrawl} (although it was discarded due to noise found in the corpus), DGT Translation Memories \cite{steinberger2012dgt}, Microsoft Translation and User Interface Strings Glossaries\footnote{https://www.microsoft.com/en-us/language/translations} from multiple releases up to 2018, the Tilde MODEL corpus \cite{Rozis2017}, WikiMatrix \cite{schwenk2019wikimatrix}, Digital Corpus of the European Parliament \cite{hajlaoui2014dcep}, JRC-Acquis \cite{Steinberger2006}, Europarl \cite{Koehn2005}, and the QCRI Educational Domain Corpus \cite{abdelali2014amara}.

\subsection{Data Filtering and Pre-Processing}

First, we filtered data using Tilde's parallel data filtering methods \cite{Pinnis2018} that allow discarding sentence pairs that are corrupted, have low content overlap, feature wrong language content, feature too high non-letter ratio, etc. The exact filter configuration is defined in the paper by \cite{Pinnis2018}.

Then, we pre-processed all data using Tilde's parallel data pre-processing workflow that normalizes punctuation (quotation marks, apostrophes, decodes HTML entities, etc.), identifies non-translatable entities and replaces them with placeholders (e.g., e-mail addresses, Web site addresses, XML tags, etc.), tokenises the text using Tilde's regular expression-based tokeniser, and applies truecasing.

\begin{table}[]
\centering
\small{
\begin{tabular}{lllll}
\toprule
\multirow{2}{*}{Scenario} & Lang. & \multirow{2}{*}{Raw} & \multicolumn{2}{c}{Filtered} \\
 & pair &  & Tilde & +DCCEF \\ \midrule
\multirow{2}{*}{(c)} & \enpl & \multirow{2}{*}{10.8M} & \multirow{2}{*}{6.5M} & 4.3M \\
 & \plen &  &  & 4.3M \\ \midrule
\multirow{2}{*}{(u)} & \enpl & \multirow{2}{*}{55.4M} & \multirow{2}{*}{31.5M} & 23.3M \\
 & \plen &  &  & 24.1M \\ \midrule
\multirow{2}{*}{(u) w/o PC} & \enpl & \multirow{2}{*}{48.8M} & \multirow{2}{*}{27.0M} & 21.6M \\
 & \plen &  &  & 21.3M \\ \bottomrule
\end{tabular}
}
\caption{Parallel data statistics before and after filtering. (c) - constrained, (u) - unconstrained, ``w/o PC'' - ``without ParaCrawl''.}
\label{tab:data-parallel}
\end{table}

In preliminary experiments, we identified also that morphology-driven word splitting \cite{Pinnis2017} for English$\leftrightarrow$Polish allowed us to increase translation quality by approximately 1 BLEU point. The finding complies with our findings from previous years \cite{Pinnis2018news,pinnis-EtAl:2017:WMT}. Therefore, we applied morphology-driven word splitting also for this year's experiments.

Then, we trained baseline NMT models (see Section~\ref{sec:baseline}) and language models, which are necessary for dual conditional cross-entropy filtering (DCCEF) \cite{junczys2018dual} in order to select parallel data that is more similar to the news domain (for usefulness of DCCEF, refer to Section~\ref{sec:dccef}). For in-domain (i.e., news) and out-of-domain language model training, we used four monolingual datasets of 3.7M and 10.6M sentences\footnote{The sizes correspond to the smallest monolingual in-domain dataset, which in both cases were news in Polish. For other datasets, random sub-sets were selected.} for the constrained and unconstrained scenarios respectively. Once the models were trained, We filtered parallel data using DCCEF. The parallel data statistics before and after filtering are given in Table~\ref{tab:data-parallel}.

For our final systems, we also generated synthetic data by randomly replacing one to three content words on both source and target sides with unknown token identifiers. This has shown to increase robustness of NMT systems when dealing with rare or unknown phenomena \cite{Pinnis2017}. This process almost doubles the size of the corpora, therefore, this was not done for the datasets that were used for the experiments documented in Section~\ref{sec:exp}.

For backtranslation experiments, we used all available monolingual data from the WMT shared task on news translation. In order to make use of the Polish CommonCrawl corpus, we scored sentences using the in-domain language models and selected top-scoring sentences as additional monolingual data for back-translation.

Many of the data processing steps were sped up via parallelization with GNU Parallel \cite{Tange2011a}.

\section{Experiments}
\label{sec:exp}

In this section, we describe the details of the methods used and experiments performed to identify the best-performing recipes for training of Tilde's NMT systems for the WMT 2020 shared task on news translation. All experiments that are described in this section were carried out on the
constrained datasets unless specifically indicated that also unconstrained datasets were used.

\subsection{NMT architecture}
\label{sec:arch}
All NMT systems that are described further have the Transformer architecture
\cite{vaswani2017attention}. We trained the systems using the Marian toolkit
\cite{mariannmt}. The Transformer \textit{base} model configuration was used
throughout the experiments except for the experiments with the \textit{big}
model configuration that are described in Section~\ref{sec:res}. We used
gradient accumulation over multiple physical batches (the
\textit{-\--optimizer-delay} parameter in Marian) to increase the effective batch
size to around 1600 sentences in the \textit{base} model experiments and 1000
sentences in \textit{big} model experiments. The Adam optimizer with a learning
rate of 0.0005 and with 1600 warm-up update steps (i.e., the learning rate linearly rises during
warm-up; afterwards decays proportionally to the inverse of the square root of
the step number) was used. For language model training, a learning rate of 0.0003 was used.

\subsection{Baseline models}
\label{sec:baseline}

We trained baseline models using the Transformer \textit{base} configuration as defined in Section~\ref{sec:arch}. The validation results for the baseline NMT systems are provided in Table~\ref{tab:dccef}. As we noticed last year that the ParaCrawl corpus contained a large proportion (by our estimates up to 50\%) \cite{pinnis-krilauks-rikters:2019:WMT} of machine translated content, we trained baseline systems with and without ParaCrawl. It can be seen that when training the \enpl ~unconstrained system using ParaCrawl, we loose over 2 BLEU points. This is because most machine translated content is on the non-English (in this case Polish) side. For the \plen\ direction, the machine-translated content acts as back-translated data and, therefore, does not result in quality degradation. Further, our \plen\ systems are trained using ParaCrawl, and \enpl\ systems -- without ParaCrawl.

\subsection{Dual Conditional Cross-Entropy Filtering}
\label{sec:dccef}

After the baseline systems, we analysed whether DCCEF allows improving translation quality. The validation results in Table~\ref{tab:dccef} show that translation quality increases for the constrained systems, but degrades for the unconstrained systems. Further, we used DCCEF only for the constrained scenario systems.

\begin{table}[]
\centering
\begin{tabular}{lrr}
\toprule
System & \multicolumn{1}{l}{\enpl} & \multicolumn{1}{l}{\plen} \\ \midrule
\multicolumn{3}{l}{\textbf{Constrained}} \\
~Baseline & 21.67 & 32.69 \\
~+DCCEF & \textbf{22.19} & \textbf{33.45} \\ \midrule
\multicolumn{3}{l}{\textbf{Unconstrained}} \\
~Baseline & 21.86 & \textbf{33.08} \\
~+DCCEF & 22.51 & 30.86 \\
~Baseline w/o ParaCrawl & \textbf{24.29} & 29.47 \\
~+DCCEF & 22.60 & 28.59 \\ \bottomrule
\end{tabular}
\caption{Comparison of baseline NMT systems trained on data that were prepared with and without DCCEF.}
\label{tab:dccef}
\end{table}

\subsection{Back-translation}
\label{sec:bt}
We used monolingual data back-translation to adapt the NMT systems to the news
domain. \citet{edunov_understanding_2018} has shown that using output sampling instead of
beam-search during back-translation yields better-performing NMT systems.
Hence, we exclusively used output sampling for monolingual data
back-translation. However, due to the abundance of monolingual data for both translation directions, we experimented with different rates of upsampling and
back-translated data cutoff to
determine whether translation performance doesn't degrade in the presence of a
too small proportion of bitext data during training.

Another dimension of inquiry was with different strategies for data filtering in the preparation
of the back-translated data. \citet{ng_facebook_2019} have described a method for
domain data extraction from general domain monolingual data using domain and
out-of-domain language models. We compared said method with a simpler
alternative of using only an in-domain language model for in-domain data
scoring. We sorted the monolingual data according to the scores produced by the
in-domain language model or by the combination of in-domain and out-of-domain
language model scores and experimented with different cutoff points when
selecting data for back-translation. 

Considering the above, we carried out experiments along two dimensions -- 1)
monolingual data selection strategy, which was either \textit{combined} or
\textit{in-domain}, signifying whether the combined score of both language
models or just the score from the in-domain language model was used,
respectively, and 2) the bitext and synthetic data mixture selection strategy,
which was one of:
\begin{itemize}
\item \textit{original ratio} -- all available bitext data for the translation
  direction were combined with all back-translated data having a score $\ge 0$,
  when using the \textit{combined} selection strategy, or N top-scoring
  back-translated sentences, when using the \textit{in-domain} selection
  strategy, where N was selected to match the amount of synthetic data selected in
  the \textit{combined} case.
\item \textit{upsampled 1:1} -- the same amount of synthetic data was selected
  as with the \textit{original ratio} data mixture selection strategy, but
  bitext was upsampled to match the amount of synthetic data.
\item \textit{cutoff \{1:1, 1:2, 1:3\}} -- all available bitext data for the
  translation direction were combined with N top-scoring back-translated
  sentences where N was chosen so that the ratio of bitext to synthetic data was
  either 1:1, 1:2 or 1:3.
\end{itemize}

As a result of the above, we ended up with 96.8M sentences (14\% retained) from
the English monolingual corpus and 137M (99\% retained) sentences from the Polish
monolingual corpus after applying the \textit{combined} data selection strategy.
Consequently, the same amount of data was selected for the \textit{in-domain}
data selection strategy in the case of \textit{original ratio} and
\textit{upsampled 1:1} data mixture selection strategies (i.e. when not doing
\textit{cutoff}).

The results for back-translation experiments are presented in
Table~\ref{table:bt}. The systems use the DCCEF-filtered constrained datasets and
therefore are directly comparable to the constrained DCCEF systems in
Table~\ref{tab:dccef}.

For our final systems, we use the \textit{combined} selection strategy for \plen\ and the \textit{in-domain} selection strategy for \enpl. For unconstrained systems, we identified that there is no significant difference between translation quality; we used the \textit{combined} selection strategy for both language pairs.

\begin{table}[]
  \begin{center}
    \small{
    \begin{tabular}{lccccc}
      \toprule
      \multirow{2}{*}{} & \multirow{2}{*}{\begin{tabular}[c]{@{}c@{}}orig.\\ ratio\end{tabular}} & \multirow{2}{*}{\begin{tabular}[c]{@{}c@{}}ups.\\ 1:1\end{tabular}} & \multicolumn{3}{c}{cutoff} \\ \cmidrule{4-6} 
                &       &       & 1:1   & 1:2            & 1:3   \\ \midrule
      \multicolumn{6}{l}{\textbf{\enpl}} \\
      combined  & 23.35 & 24.01 & 24.52 & 24.72          & -     \\
      in-domain & 22.10 & 22.92 & 25.02 & \textbf{25.28} & 25.24 \\ \midrule
      \multicolumn{6}{l}{\textbf{\plen}} \\
      combined  & 31.19 & 33.45 & 33.29 & \textbf{33.60} & -     \\
      in-domain & 29.67 & -     & 33.40 & 33.28          & -     \\ \bottomrule
    \end{tabular}
    }
    \caption{\label{table:bt} Back-translation experiment results. }
  \end{center}
\end{table}

\subsection{QHAdam optimizer}
Last year \cite{pinnis-krilauks-rikters:2019:WMT} we used the QHAdam optimizer
\cite{ma_quasi-hyperbolic_2018} for model training, however, we hadn't treated
QHAdam and Adam the same in the experimental process, having dedicated
substantially more effort to optimizer hyper-parameter tuning for QHAdam than
Adam. To make an unbiased comparison of the two optimizers, we trained
corresponding system variants using QHAdam for the \textit{combined cutoff 1:2},
\textit{in-domain cutoff 1:2} and \textit{in-domain cutoff 1:3} systems from
Section~\ref{sec:bt} in the \enpl\ translation direction. The BLEU scores for
the experiments are found in Table~\ref{table:qhadam}. We see that QHAdam
performs no better than Adam. We had also done more extensive experiments
comparing QHAdam to Adam for a range of learning rate and warm-up step parameter
settings on a different dataset, which showed a similar trend, however we do not
present those results here. As a result, we didn't choose QHAdam over Adam in
this year's competition.

We note, however, that we used the recommended safe defaults for the QHAdam's
hyperparameters -- $v_1 = 0.8, v_2 = 0.7$ -- and we haven't performed a search
over these values which could have yielded different results.

\begin{table}[]
  \centering
    \begin{tabular}{lccc}
      \toprule
      \multirow{2}{*}{} & \multirow{2}{*}{\begin{tabular}[c]{@{}c@{}}combined\\ cutoff 1:2\end{tabular}} & \multicolumn{2}{c}{in-domain cutoff} \\ \cmidrule{3-4} 
             &                & 1:2            & 1:3            \\ \midrule
      Adam   & 24.72          & \textbf{25.28} & \textbf{25.24} \\
      QHAdam & \textbf{24.86} & 24.98          & 25.00          \\ \bottomrule
    \end{tabular}
    \caption{ BLEU scores for the QHAdam experiments in the
      \enpl\ translation direction. }
    \label{table:qhadam}
\end{table}

\subsection{Right-to-Left Re-Ranking}
\label{sec:r2l}
\citet{morishita_ntts_2018} report improving the translation performance by
using right-to-left (R2L) re-ranking. The method employs a right-to-left model
to re-score the $n$-best list outputs of a regular -- left-to-right -- model by
multiplying both models' translation probabilities. We implement R2L re-ranking
the same as \citet{morishita_ntts_2018}, but opted to use $n$-best lists with
$n = 12$ (instead of $n = 10$).

The R2L re-ranking experiments were performed during the final stages of the
competition, hence the baseline systems for those experiments were the final
systems that were being prepared for submission to the news translation task.
Therefore we present the results in Table~\ref{tab:results} in the Results
section. We find similar improvements as \citet{morishita_ntts_2018}, albeit
they are slightly smaller.

\section{Final Systems}
\label{sec:res}
We chose the best performing system variants from Section~\ref{sec:exp} to serve
as a base for the final submission for the news translation task. For the constrained scenario, we trained final systems using parallel data that were filtered with Tilde's filtering methods and DCCEF, back-translated monolingual data using a ratio of 1:2 (different data selection methods were applied for both translation directions), and synthetic data featuring unknown phenomena. For the unconstrained scenario, we trained final systems using parallel data that were filtered only with Tilde's filtering methods, back-translated monolingual data that were selected using the \textit{combined} data selection strategy using a ratio of 1:1, and synthetic data featuring unknown phenomena. All models were trained using the Adam optimiser. 

When preparing the final systems,
we also employed R2L re-ranking (see Section~\ref{sec:r2l}), ensembling of the best three models,
and trained Transformer models using the \textit{big} model configuration.

\section{Results}
\label{sec:res}

The BLEU scores for the systems that were evaluated for the final submission are
shown in Table~\ref{tab:results}. The results show that right-to-left reranking increased translation quality for all systems. For the \enpl\ translation direction, the best results were achieved when using ensembles of three models and better results were achieved by the unconstrained systems. However, for the \plen\ translation direction, the unconstrained systems achieved lower results than the constrained systems. The best results were achieved by the Transformer big model; ensembling did not improve results.

In overall, the results differ from what we have observed in previous years. Back-translation for \plen\ did not improve results, which raises a question of a possible domain mismatch between the monolingual data we back-translated and the development data. Unconstrained systems are only slightly better than constrained systems for \enpl\ and even subpar for the \plen\ translation direction, which shows that current NMT methods are not able to benefit from larger datasets. Hence, having in-domain data is more important.

\begin{table}[]
\centering
\begin{tabular}{lrr}
\toprule
 & Constrained & Unconstrained \\ \midrule
\multicolumn{3}{l}{\textbf{\plen}} \\ 
Base & 33.48 & 32.63 \\
+R2L & 34.34 & 33.29 \\
Big & 33.79 & 33.15 \\
+R2L & \underline{\textbf{34.83}} & 33.45 \\
Ensemble of 3 & 34.19 & 33.39 \\
+R2L & 34.80 & 33.53 \\ \midrule
\multicolumn{3}{l}{\textbf{\enpl}} \\
Base & 25.64 & 26.12 \\
+R2L & 26.24 & 26.52 \\
Big & 25.59 & 26.47 \\
+R2L & 26.70 & 26.78 \\
Ensemble of 3 & 26.07 & 26.86 \\
+R2L & \underline{26.73} & \underline{\textbf{27.12}} \\ \bottomrule
\end{tabular}
\caption{Final system evaluation results (BLEU scores) on validation data (bold marks best scores; submitted systems are underlined).}
\label{tab:results}
\end{table}

\section{Conclusion}
\label{sec:conclusion}

In this paper, we described Tilde's NMT systems for the WMT shared task on news translation. This year, we trained constrained and unconstrained systems for the English$\leftrightarrow$Polish language pair. We detailed the methods applied and the training recipes.

During our experiments, we identified several avenues of possible further research. We saw that larger datasets even after applying data selection methods did not allow improving translation quality (at least not significantly). We made a similar observation also previous years when participating in WMT. We saw in our results also that back-trannslation did not yield positive results for \enpl. We hypothesise that there may be a domain mismatch between the data we used for training and the newsdev2020 dataset. However, this requires further investigation. 

\section*{Acknowledgements} \label{sec-acknowledgements}

The research has been supported by the European Regional Development Fund
  within the research project ``Multilingual Artificial Intelligence Based Human Computer Interaction'' No. 1.1.1.1/18/A/148. We thank the High Performance Computing Center of Riga Technical University for providing access to their GPU computing infrastructure.

\bibliographystyle{acl_natbib}
\bibliography{anthology,emnlp2020}

\end{document}